%% file: main.tex
\documentclass[10pt,twocolumn,letterpaper]{article}
\usepackage{indentfirst}
\usepackage[pagenumbers]{cvpr} 


\definecolor{cvprblue}{rgb}{0.21,0.49,0.74}
\usepackage[pagebackref,breaklinks,colorlinks,allcolors=cvprblue]{hyperref}

\def\paperID{7455} 
\def\confName{CVPR}
\def\confYear{2025}

\title{One Pixel is All I Need}

\author{Deng Siqin\\
Hainan University\\
58 Renmin Street, Haikou, Hainan\\
{\tt\small 22210839000016@hainanu.edu.cn}
\and
Zhou Xiaoyi\\
Hainan University\\
58 Renmin Street, Haikou, Hainan\\
{\tt\small xy.zhou@hainanu.edu.cn}
}

\begin{document}
\maketitle

\input{sec/0_abstract}
\input{sec/1_intro}
\input{sec/2_Related_Works}
\input{sec/3_finalcopy}
\input{sec/4_The_Proposed_WorstViT_Framework}
\input{sec/5_Conclusion}
{
    \small
    \bibliographystyle{ieeenat_fullname}
    \bibliography{main}
}

\input{sec/X_suppl}

\end{document}

%% file: sec/0_abstract.tex
\begin{abstract}
    \indent Since their introduction, Vision Transformers (ViTs) have demonstrated record-breaking performance in various visual tasks. However, their robustness against potential malicious attacks has also attracted increasing attention from researchers. Backdoor attacks involve forming a strong association between a specific trigger and a target label. The backdoored model correctly identifies clean images but predicts the attacker-specified label when the trigger is present in the input. Previous work has attempted to use triggers that are different but similar to those used during training. We refer to such patterns that can still trigger the backdoor despite being different from the original trigger as quasi-triggers. We found that, compared to CNNs, ViTs exhibit higher attack success rates for quasi-triggers. Additionally, when some backdoor features are present in clean samples, the backdoor may be suppressed in some inputs, causing the original trigger to fail, while quasi-triggers can successfully trigger the backdoor and complete the attack.
    The perturbation sensitivity distribution map (PSDM) of the model is generated by computing and summing gradients over a large number of inputs. This map shows the model's sensitivity to content-agnostic perturbations in the input. Utilizing the PSDM can help us better alter the model's decisions, an angle that past malicious attacks have not considered. Notably, the PSDM of ViTs exhibits a clear patch-like pattern, where the central pixels of each patch are more sensitive to perturbations than the edges. The PSDM is specifically used to guide the use of quasi-triggers.
    Based on these findings, we designed a simple yet effective data poisoning backdoor and its variants for ViT models. We only need an extremely low poisoning rate, train for one epoch, and modify a single pixel to successfully attack all validation images. We name this attack "WorstVIT."
\end{abstract}

%% file: sec/1_intro.tex
\section{Introduction}
\label{sec:intro}
The Transformer \cite{vaswani2017attention} was introduced to address sequence transduction problems in machine translation tasks \cite{bahdanau2014neural}. By leveraging its self-attention mechanism, the Transformer effectively solved long-distance dependency issues, achieving significant success in natural language processing tasks \cite{devlin2018bert, yenduri2024gpt, liu2019roberta, lan2019albert, zhou2021informer}. This success sparked interest among researchers in applying the architecture to visual tasks. The most well-known application, Vision Transformers (ViTs) \cite{dosovitskiy2020image}, segments images into patches and uses the linear embeddings of these patches as input to the Transformer. Relying solely on the self-attention mechanism, ViTs have outperformed Convolutional Neural Networks (CNNs) in multiple visual tasks, including image classification \cite{li2022efficientformer, strudel2021segmenter, wang2021pyramid, liu2021swin, touvron2021training, yuan2021tokens}.

However, like CNNs, ViTs are equally vulnerable to adversarial attacks \cite{szegedy2013intriguing, goodfellow2014explaining, mahmood2021robustness, aldahdooh2021reveal, liu2022imperceptible} and backdoor attacks \cite{chen2017targeted, gu2017badnets, sun2022backdoor, saha2020hidden, bai2021targeted, gao2021backdoor}. Backdoor attacks manipulate model training to establish a strong association between triggers and target labels. After being trained with poisoned data, the Deep Neural Network (DNN) maintains high accuracy on clean inputs but produces the attacker's designated prediction when the input contains the trigger. Even if the trigger slightly differs from the one used during training, as long as it possesses certain features of the trigger, it can partially activate the backdoor; we term these quasi-triggers.

Previous research has not systematically compared the robustness of CNNs and ViTs under quasi-trigger attacks. Our work fills this gap and finds that quasi-triggers achieve much higher attack success rates on ViT models compared to CNN models. Even many samples that fail with the original trigger can be successfully attacked using quasi-triggers. By comparing the original scores of these failed samples, we found that this is because, in some clean samples that possess partial trigger features but maintain their original labels, the model learns to treat certain features of these clean samples as backdoor suppression patterns. We can disrupt or bypass these backdoor suppression patterns by using quasi-triggers, thereby significantly increasing the attack success rate.

Past efforts to explain model decisions have involved various visualization techniques \cite{smilkov2017smoothgrad, selvaraju2017grad}, often using heatmaps to show which regions of an individual input image contribute to the model's prediction. When creating adversarial examples, gradients are typically computed for specific samples, with areas of larger gradients usually corresponding to the main content of the image. This means that altering the main content of the image can better change the model's decision, but it overlooks the model's sensitivity to content-agnostic perturbations. To explore the sensitivity distribution of the model to content-agnostic perturbations, we computed and summed gradients over a large number of images from different categories or randomly generated unordered images. The resulting gradient accumulation map reveals the sensitivity distribution of the model to content-agnostic perturbations, which we call the perturbation sensitivity distribution map (PSDM). For ViT models, this distribution exhibits a highly regular patch-like pattern, as shown in Figure~\ref{fig:VIT-Perturbation Sensitivity Distribution Maps}, where the central pixels of each patch are more sensitive than the edges. Placing the trigger at the center of a patch results in a higher attack success rate compared to placing it at the edge.

Based on these findings, this paper proposes a backdoor attack called WorstVITs that leverages the unique perturbation sensitivity distribution of ViT models. For each image in the test set, modifying just one pixel can change the model's prediction to the specified label. For a significant portion of inputs, altering one value among 150528 (3*224*224) channel values suffices to complete the attack. This represents the minimal alteration to the original image in the field of model security research.

The main contributions of this paper are:
\begin{figure}[htbp]
    \centering
    \begin{subfigure}{0.32\textwidth}
        \centering
        \includegraphics[width=\linewidth]{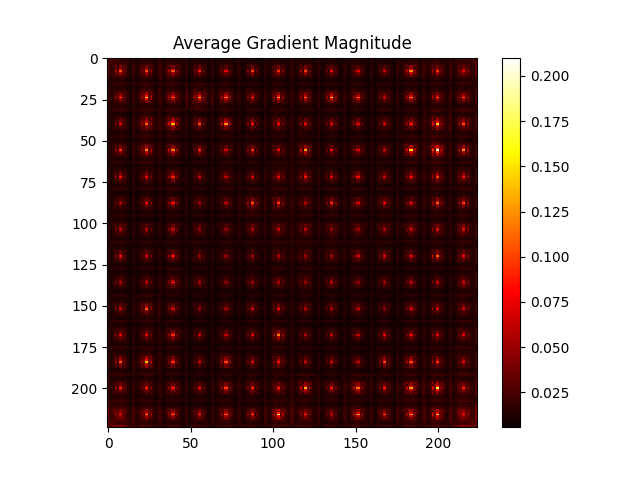}
        \caption{ViT}
        \label{fig:vitsmall_16}
    \end{subfigure}
    \hfill
    \begin{subfigure}{0.32\textwidth}
        \centering
        \includegraphics[width=\linewidth]{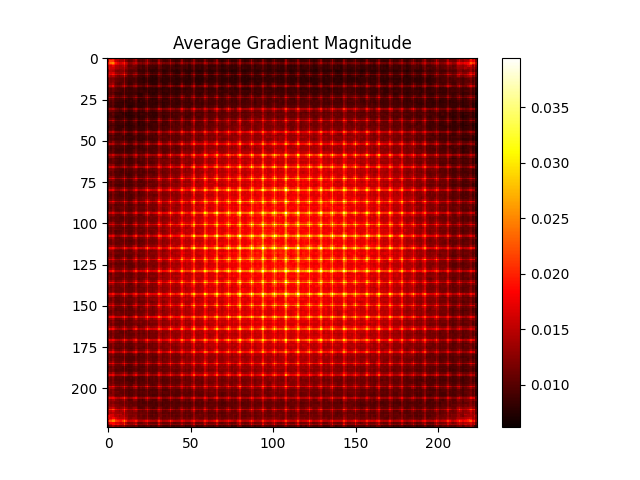}
        \caption{VGG}
        \label{fig:vgg16}
    \end{subfigure}
    \hfill
    \begin{subfigure}{0.32\textwidth}
        \centering
        \includegraphics[width=\linewidth]{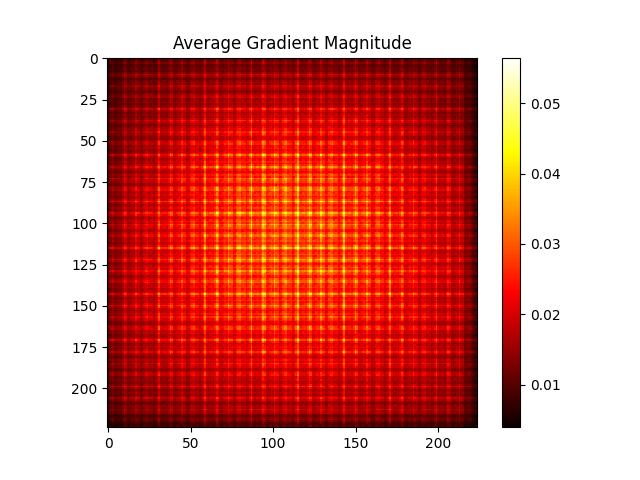}
        \caption{ResNet}
        \label{fig:resnet18}
    \end{subfigure}
    \caption{Perturbation Sensitivity Distribution Maps (PSDMs) for different configurations of models. The PSDMs exhibit a unique patch-like pattern, indicating that the model is more sensitive to perturbations at the centers of the patches compared to the edges.}
    \label{fig:VIT-Perturbation Sensitivity Distribution Maps}
\end{figure}

\begin{itemize}
    \item \textbf{Formal Introduction of Quasi-Triggers:} We discover patterns within the model that can suppress the activation of backdoors by triggers, which we refer to as backdoor suppression patterns. We formally introduce the concept of quasi-triggers and compare the differences in quasi-trigger attacks between Vision Transformers (ViTs) and Convolutional Neural Networks (CNNs). Our findings show that quasi-triggers achieve higher attack success rates in ViTs. Based on these insights, we propose a simple method to prevent the suppression of backdoor patterns and conduct a probing backdoor attack using quasi-triggers in a white-box setting. 
    \item \textbf{Perturbation Sensitivity Distribution Map (PSDM):} We introduce the PSDM, which reveals the unique patch-wise input perturbation sensitivity distribution of ViTs. The PSDM not only shows the model's sensitivity to adversarial perturbations but also exhibits the same regularities in backdoor attacks. Utilizing PSDM can help attackers better alter model decisions and provides constructive suggestions for improving model robustness.
    \item \textbf{WorstVIT and Its Variants:} We propose WorstVIT, which leverages the perturbation sensitivity distribution of ViT models during inference to maximize the activation of backdoor patterns. We also introduce two variants, WorstSwin-VIT and Hidden-WorstVIT. Extensive experiments validate the effectiveness of our methods. We demonstrate our attacks in the physical world, particularly in scenarios with fixed viewpoints and large smooth regions in the background, such as traffic surveillance and photography. By modifying a single pixel, we change the prediction of every frame in a video to our specified target. Our attacks bypass mainstream backdoor defenses.Furthermore, we discuss the challenges faced by models when learning backdoor patterns in all-to-all attacks.
\end{itemize}

This paper provides readers with a new perspective on deep models and offers constructive insights to the community for improving model robustness, especially highlighting the extreme vulnerability of ViT models.

%% file: sec/2_Related_Works.tex
\section{Related Works}
\label{sec:formatting}

\subsection{Backdoor Attacks and Defenses}

\indent In the field of deep learning, backdoor attacks are techniques that enable attackers to inject specific poisoned samples into the training data, causing the trained deep neural network (DNN) to exhibit hidden trigger patterns. These patterns lead the model to misclassify inputs containing the trigger to an attacker-specified target class. 

The BadNets attack \cite{gu2017badnets} was the first backdoor attack targeting DNNs. It involved randomly selecting some images from the dataset, adding triggers, and modifying their labels to the target label chosen by the attacker. Through this training process, the backdoored model could correctly classify benign images while misclassifying images with triggers as the designated target class \cite{chen2017targeted, gu2017badnets}. Subsequently, backdoor attacks have evolved into various sophisticated and stealthy forms \cite{saha2020hidden, bai2021targeted, gao2021backdoor}, being applied in a wide range of scenarios \cite{sun2022backdoor, yang2024not, li2024nearest}, covering extensive computer vision tasks \cite{feng2022fiba, yu2023backdoor, li2022backdoor}. Some of these attacks have achieved alarming success rates through carefully designed backdoors.

For example, Li \cite{li2022backdoor} found that backdoor attacks tend to be more successful on smaller datasets. Doan et al.'s LIRA \cite{doan2021lira} drew inspiration from GANs, integrating trigger generation and poisoning processes within a constrained optimization framework, achieving a 100\% attack success rate on small datasets. Yuan et al.'s BadViT \cite{yuan2023you} generated adversarial patch triggers capable of capturing most of the attention of Vision Transformers (ViTs), reaching a 100\% attack success rate with only 1\% poisoning rate after just one training epoch.

However, these approaches not only require access to the complete model architecture and parameters but also necessitate control over the entire training process and all the training data, making it challenging to poison models efficiently within a short time frame. While the basic assumption in backdoor attacks is that the attacker has full knowledge of the backdoor model's structure and parameters, and often controls the entire training process—implying possession of the entire dataset—in practical attacks, data poisoning tends to be easier to implement than controlling the model's training process.

Similarly, under the white-box setting, we successfully attacked all images in the ImageNet val set using quasi-triggers. For images with weaker natural features, we modified only a single value in the entire input matrix. For potentially failed attacks, since our quasi-trigger is a single pixel without a fixed value that can appear anywhere in the image, we can place multiple quasi-triggers simultaneously to launch stronger attacks. This approach differs from that of Yin \cite{yin2024enhanced}, who placed triggers at four fixed positions during training and attack, adjusting the pixel values of the triggers. Our quasi-triggers do not have fixed positions; their placement depends on the model's perturbation sensitivity distribution and the current input gradient.

As the threat of backdoor attacks becomes increasingly evident, research interest in backdoor defenses has also grown. Neural Cleanse \cite{wang2019neural} employs reverse engineering to generate triggers for backdoored models and identifies the attacked classes by comparing the anomaly scores of different classes. FinePruning \cite{liu2018fine} precisely prunes the model to remove the backdoor while maintaining the model's accuracy on clean datasets. SCALE\_UP \cite{guo2023scale} leverages the phenomenon of scaled prediction consistency, where the predictions of poisoned samples remain more consistent compared to those of benign samples when all pixel values are scaled up. It uses this property to detect and identify backdoor patterns during the prediction phase.

\subsection{Vision Transformer}

In natural language processing, the Transformer \cite{vaswani2017attention}, which relies solely on the attention mechanism, is the current dominant model architecture. Vision Transformers (ViTs) \cite{dosovitskiy2020image} apply this concept to computer vision tasks by dividing images into patches, treating these patches as tokens similar to words in natural language. The Transformer is then applied to these patches. Later, to address some of the issues with ViTs \cite{chen2021visformer, xiao2021early, han2021transformer}, improved versions \cite{touvron2021training, liu2021swin, heo2021rethinking, lou2023transxnet} were proposed. However, the core idea of treating images as sequences of patches and processing them through the attention mechanism remains central to ViTs and their variants. Specifically, for each patch in the input, the same set of convolutional kernels or fully connected layer parameters is reused. We believe this is the reason why the perturbation sensitivity distribution in ViTs exhibits a patch-wise pattern.

\subsection{Visualization Methods for Deep Models}
For deep models handling computer vision tasks, many techniques have been proposed to explain and visualize model decisions \cite{smilkov2017smoothgrad, selvaraju2017grad}. These techniques typically highlight which regions of the input image are most important for the model's decision-making process. However, there has been little research on the sensitivity of models to different regions of the input itself, independent of the specific content of the image. To fill this gap, we propose the Perturbation Sensitivity Distribution Map (PSDM). The PSDM reveals the sensitivity distribution of the model to content-agnostic perturbations in the input, showing that the model pays varying degrees of attention to perturbations in different regions of the input. Particularly, in Vision Transformers (ViTs), this sensitivity distribution exhibits a clear patch-wise pattern, which is crucial for the implementation of our attack.

\subsection{One pixel attack}

In the field of adversarial attacks, Su et al. \cite{su2019one} proposed a differential evolution-based method to generate single-pixel adversarial perturbations. Their method achieved success rates of 67.9\% on the 32x32 CIFAR-10 dataset and 16.04\% on the 224x224 ImageNet dataset by modifying one pixel of the images. Unlike their work, ours focuses on backdoor attacks rather than adversarial examples. They define success as the model output differing from the original label, whereas we require the model's prediction to be modified to a specific target label.

Another relevant work is by Li et al. \cite{li2020rethinking}, who discussed using a single pixel for data poisoning attacks in CIFAR-10. However, their primary focus was on using outlier detection algorithms to remove poisoned inputs, and the use of a single pixel was primarily illustrative. Our attack leverages the model's perturbation sensitivity distribution and strategically places triggers to avoid conflicts with clean data. As a result, our attack often does not degrade the model's accuracy on clean tasks.

%% file: sec/3_finalcopy.tex
\section{Inspirations and Phenomenon}

\begin{figure*}[ht]
    \centering
    \begin{subfigure}{0.32\textwidth}
        \centering
        \includegraphics[width=\linewidth]{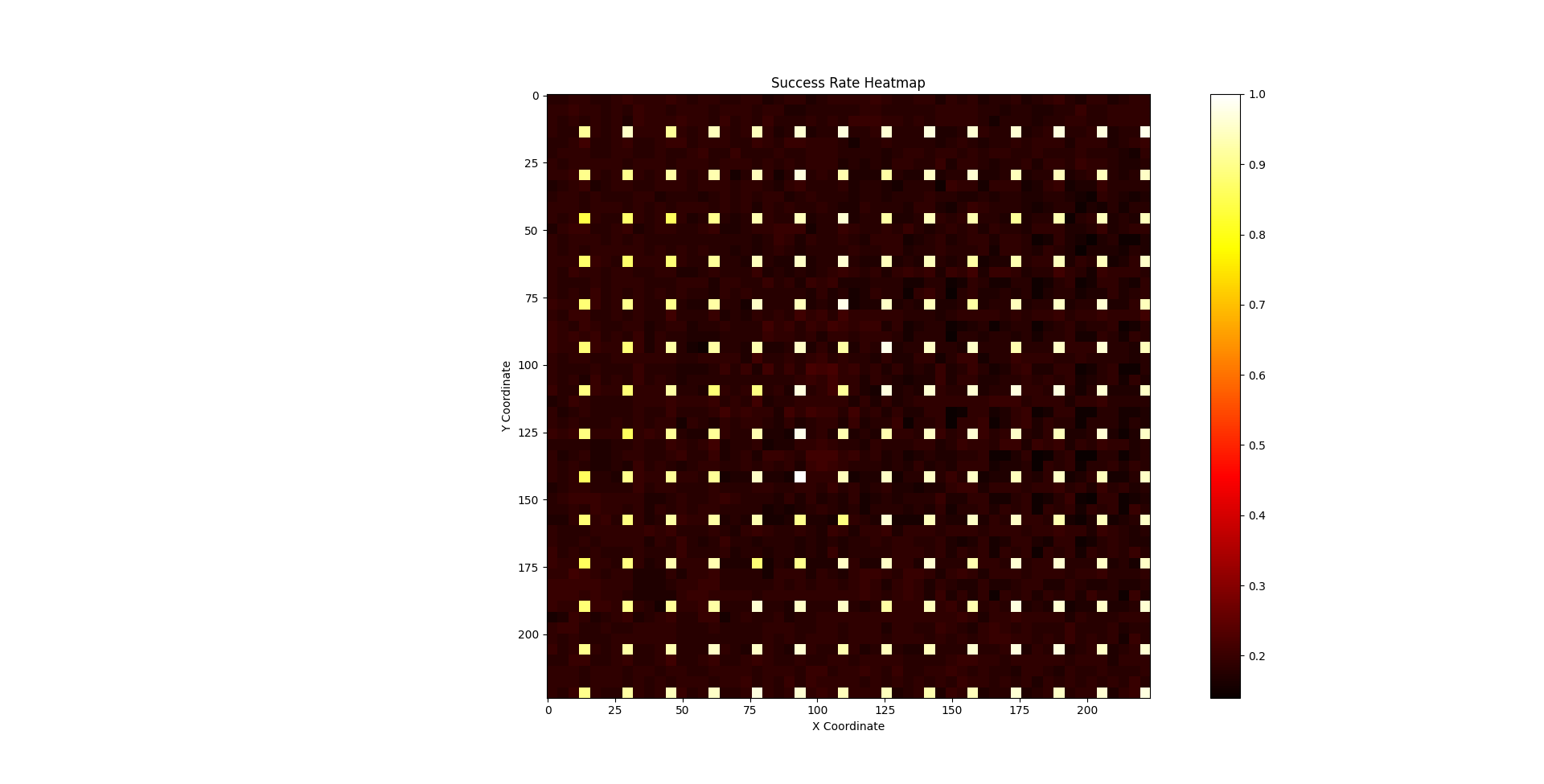}
        \caption{ViT}
        \label{fig:deit_block4_traverse_100}
    \end{subfigure}\hfill
    \begin{subfigure}{0.32\textwidth}
        \centering
        \includegraphics[width=\linewidth]{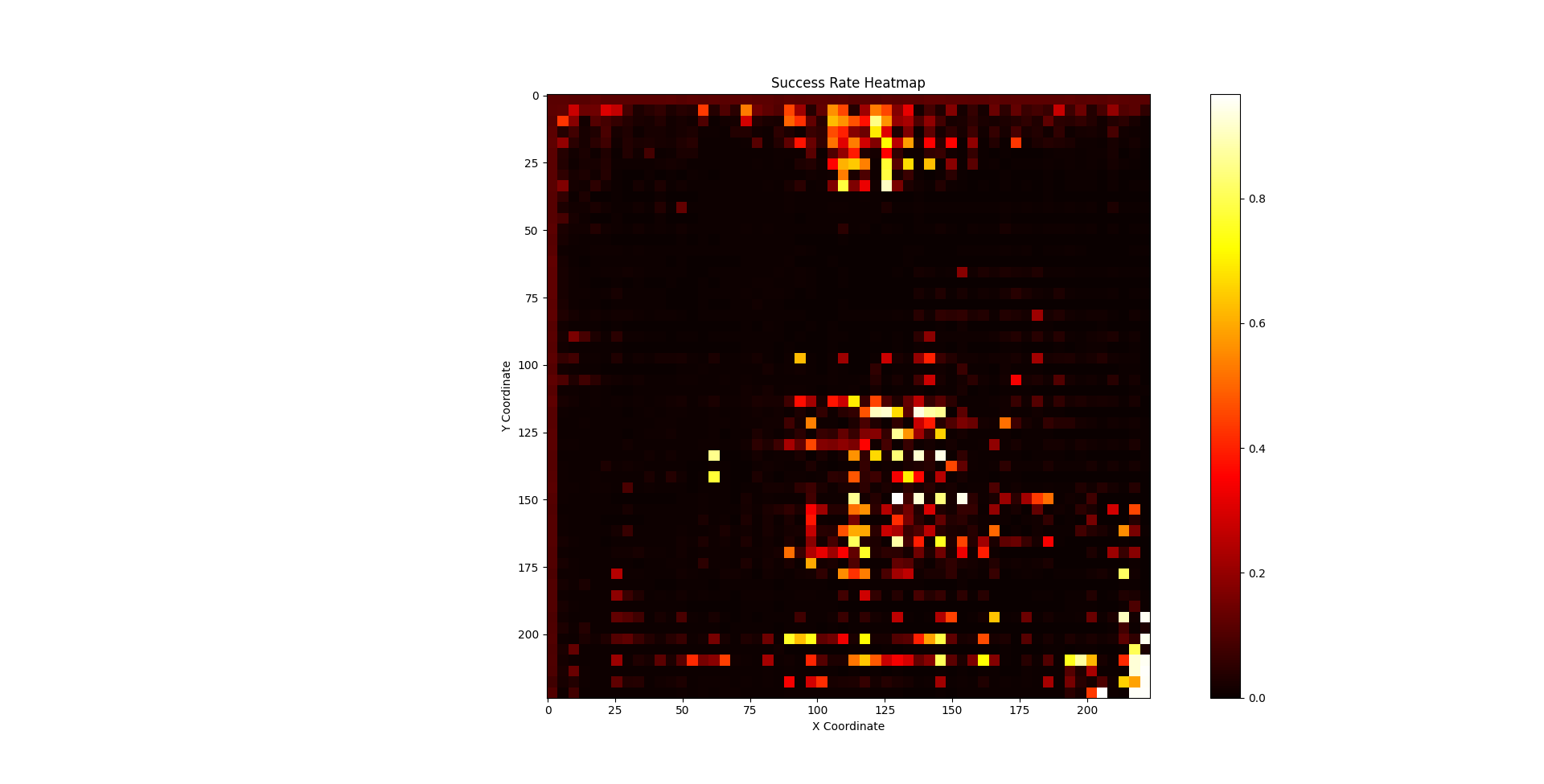}
        \caption{VGG}
        \label{fig:VGG16_block4_traverse1000}
    \end{subfigure}\hfill
    \begin{subfigure}{0.32\textwidth}
        \centering
        \includegraphics[width=\linewidth]{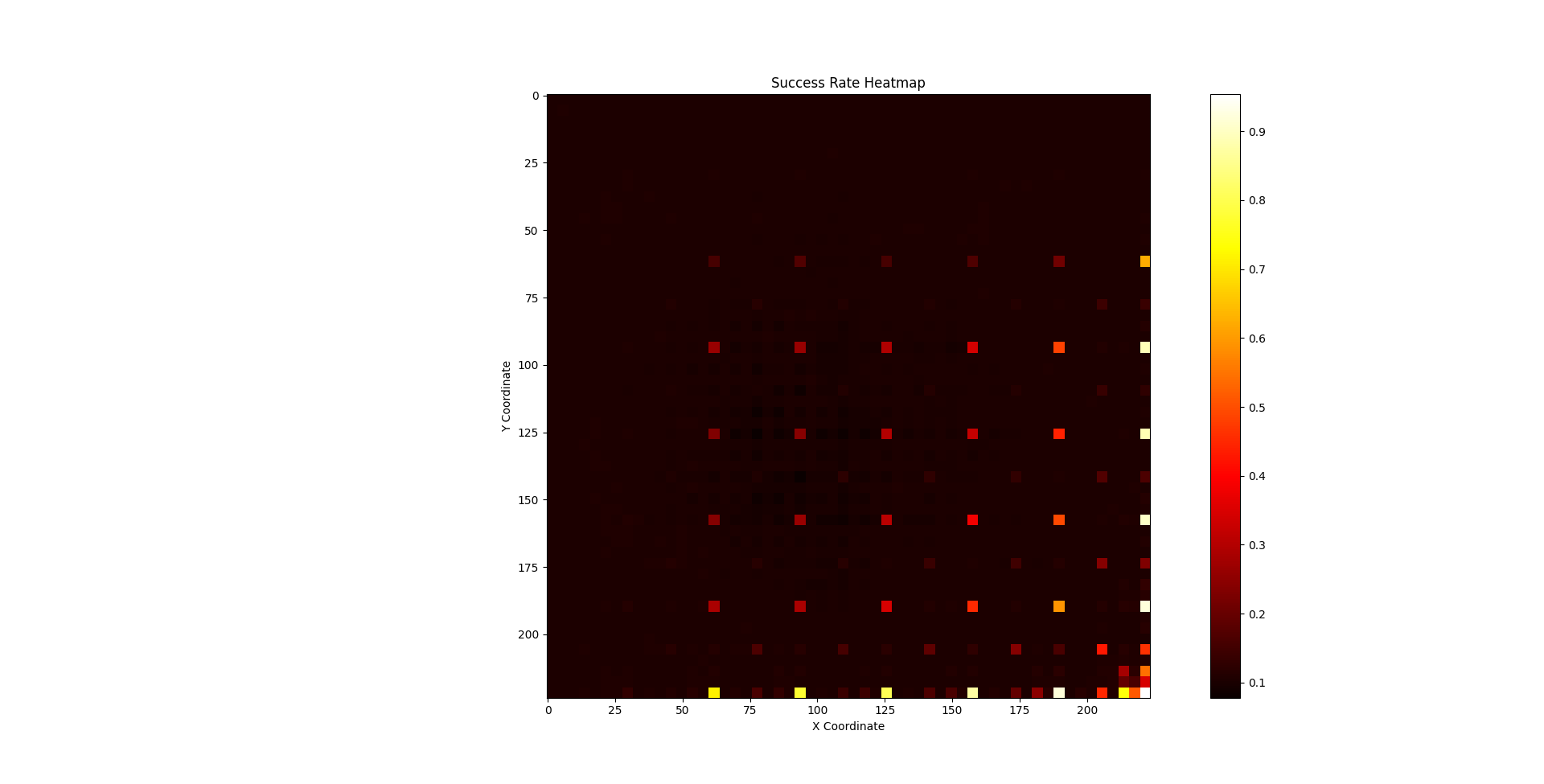}
        \caption{ResNet}
        \label{fig:resnet18_block4_traverse1000}
    \end{subfigure}
    \caption{Quasi-triggers in Vision Transformers (ViTs) exhibit good transferability across different patches.}
    \label{fig:quasi_triggers}
\end{figure*}

\begin{figure}[ht]
    \centering
    \includegraphics[width=0.32\textwidth]{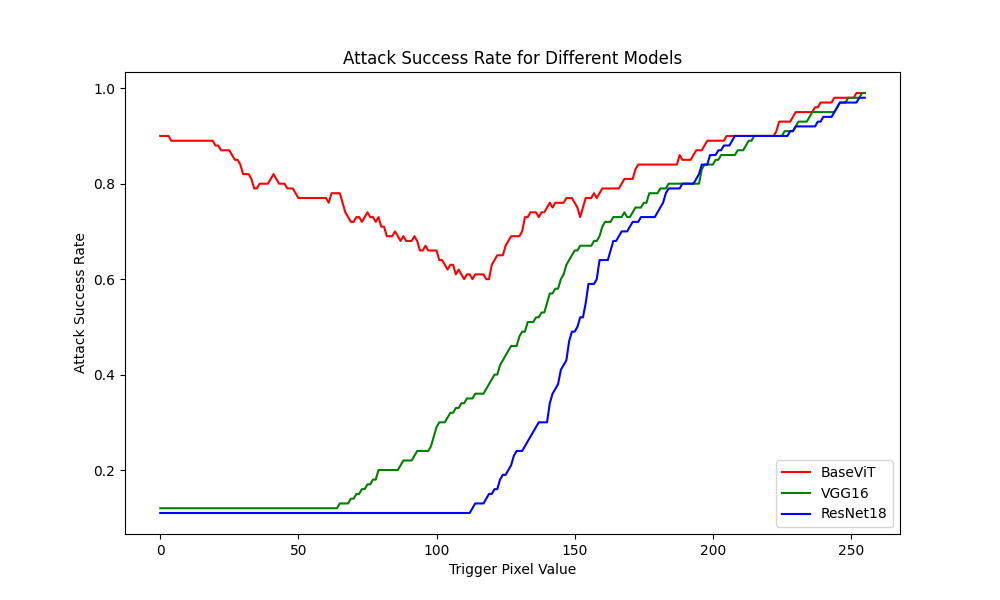} 
    \caption{Attack Success Rate for Differrent Models.}
    \label{fig:Attack Success Rate for Differrent Models}
\end{figure}

In this section, we experimentally reveal two phenomena: 

1. The unique patch-like gradient distribution characteristic of Vision Transformer (ViT) models.

2. The backdoor suppression patterns, which occurs due to the presence of partial backdoor features in clean images during backdoor attacks.

Additionally, we designed a series of experiments to compare the robustness of ViT models and Convolutional Neural Networks (CNNs) under quasi-trigger attacks.

\subsection{Perturbation Sensitivity Distribution Maps}

In some studies on adversarial examples for Vision Transformer (ViT) models \cite{yang2024not, benz2021adversarial}, adversarial samples for ViT models consistently exhibit noticeable patch-like stripes. Due to the unique working mechanism of ViT, it is reasonable that its universal adversarial perturbations (UAPs) appear in a patch-like pattern. Different textures within each patch can cause the human eye to distinguish the contours of each patch. However, when we attempted to create adversarial examples in ViT models, we still observed visible patch contours. Our hypothesis is that in ViT models, the gradients of pixels at the edges of patches differ from those at the centers of patches. Therefore, during gradient ascent on the input, patch-like stripes gradually emerge and become visible to the human eye.

By computing and summing the gradients of a large number of meaningful images and random noise images separately, we obtained the perturbation sensitivity distribution map of the model. This map reveals a content-agnostic gradient distribution, indicating that the model's sensitivity to perturbations varies across different regions of the input. Specific experimental details are provided in the Appendix.

We used a large number of open-source models and formed two groups: one with 1000 randomly selected content-rich images and another with 1000 randomly generated meaningless images. For each group, we computed the gradients for all input images and summed them. The resulting images, which we call Perturbation Sensitivity Distribution Maps (PSDMs), illustrate the content-independent sensitivity of the model to input perturbations.

As shown in the Figure~\ref{fig:VIT-Perturbation Sensitivity Distribution Maps}, ViT models consistently exhibit a clear patch-like pattern. This is understandable because the patch-embedding layer in ViT models shares the same parameters across each patch of the input, leading to similar distributions within each patch. According to the definition of gradients, ViT models are more sensitive to perturbations at the centers of patches compared to the edges.

Furthermore, we hypothesize that this pattern not only applies to adversarial examples but also to backdoor attacks. To further validate our hypothesis, we placed white square backdoors at both the edges and centers of patches. We conducted this experiment ten times with different patches, and the trigger attack success rate was consistently higher for backdoors placed at the center of the patches compared to those at the edges. Details are given in Appendix.

\subsection{Backdoor Suppression Feature}

In past backdoor attack studies, few have observed and analyzed images where the attack failed. We conducted two sets of experiments, using 4x4 white squares and 4x4 black squares separately as triggers, and found that failed attack samples can be categorized into two types:

1. High Original Scores: The true label of the image originally had a very high score, so even after the backdoor was triggered, the score for the backdoor label did not exceed the score for the true label. This situation, where the association between the trigger and the target label is insufficient, is not discussed in this paper. We focus primarily on the following scenario.
2. Backdoor Suppression: Although the trigger is present, the score for the backdoor label remains low.Similar to the work by Liu \cite{liu2023beating}, besides the backdoor trigger patterns, there is also a backdoor suppression patterns. They manually designed samples that contained both backdoor triggers and backdoor suppressors. When only the backdoor trigger was present in the image, it would be classified as the attacker-specified target class. However, when both the backdoor trigger and suppressor were present, the image would be classified as a non-target class.Unlike their work, we did not manually design suppressors. Due to the widespread presence of quasi-triggers, many images already contain partial features of backdoor triggers that correspond to the correct labels. Our intuition is that the conflict between clean samples and quasi-triggers can lead to the trigger failing to function effectively in some samples with suppression features.

To validate our hypothesis, we added borders around the trigger using both the same and opposite colors, which significantly increased the attack success rate.

This means that the trigger pattern is not limited to the pixels where the trigger is located but is formed in conjunction with other pixels. Furthermore, we can improve the attack success rate of the trigger by destroying or avoiding the suppression mode.

For more detailed information on the experiments and formulas, please refer to the appendix.

\subsection{Quasi-Triggers Robustness Comparison}
Li's previous work \cite{li2020rethinking} tested the robustness of quasi-triggers in CNN models, showing that the attack success rate decreases when the pixel values or positions of the quasi-triggers differ from those of the actual triggers. We referenced their experimental setup and placed the trigger in the bottom-right corner of the input. To facilitate the traversal of all coordinates and pixel values in the input, we set the trigger as a 4x4 white square. We conducted experiments by traversing all possible positions and pixel values of the trigger.

First, while keeping the appearance of the trigger unchanged, we altered its position during inference. We trained the backdoor by placing the trigger in the bottom-right corner of the image, which corresponds to the bottom-right corner of the last patch in the ViT model. As shown in Figure~\ref{fig:quasi_triggers}, compared to CNN models, the trigger in ViT models can effectively migrate to the same position in other patches. Quasi-triggers not only exhibit a highly regular pattern but also achieve a higher attack success rate.

Next, while keeping the position of the trigger unchanged, we calculated the attack success rate for pixel values ranging from 0 to 255. As shown in the Figure~\ref{fig:Attack Success Rate for Differrent Models}, in ViT models, the attack success rate of quasi-triggers with various pixel values is consistently higher than in CNN models. Additionally, in ViT models, even when we trained with a white trigger, a black trigger also shows a high attack success rate.

In the previous section, to disrupt the backdoor suppression pattern, we added a border to the trigger. In this section, when the original trigger fails to attack, simply modifying the pixel values within the trigger (e.g., using black pixels) can succeed without altering other pixels. Considering that convolutional kernels with strides smaller than their size naturally smooth feature maps, we hypothesize that ViT models may be more sensitive to the contrast of triggers. Therefore, in Section 4, we use the contrast between a single-pixel pattern and other pixels within the same patch as the trigger. The modified pixel and its patch together form our backdoor trigger pattern.

%% file: sec/4_The_Proposed_WorstViT_Framework.tex
\section{The Proposed WorstViT Framework}

In the previous section, we compared the robustness of quasi-triggers in CNN and ViT models. In ViT, quasi-triggers exhibit the following three characteristics:

\begin{figure*}[htbp]
    \centering
    \begin{subfigure}{0.32\textwidth}
        \centering
        \includegraphics[width=\linewidth]{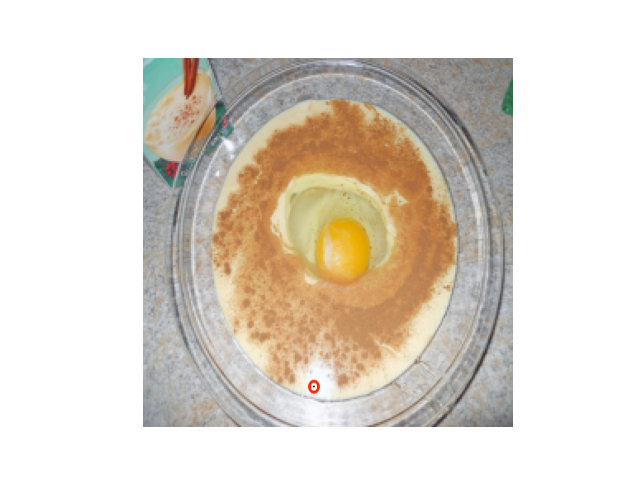}
        \caption{Example 1}
        \label{fig:Figure_1}
    \end{subfigure}\hfill
    \begin{subfigure}{0.32\textwidth}
        \centering
        \includegraphics[width=\linewidth]{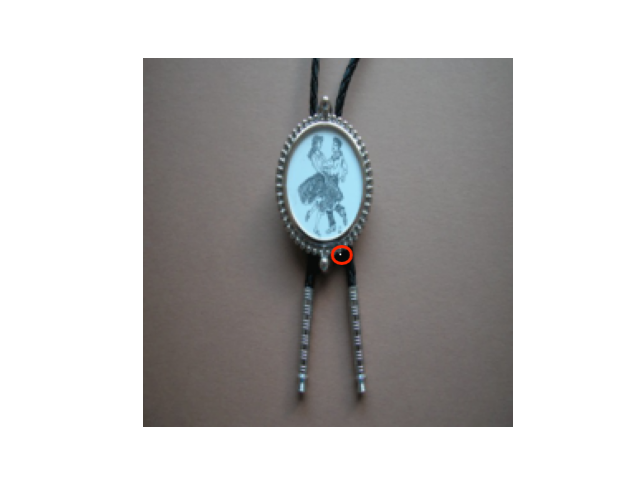}
        \caption{Example 2}
        \label{fig:Figure_2}
    \end{subfigure}\hfill
    \begin{subfigure}{0.32\textwidth}
        \centering
        \includegraphics[width=\linewidth]{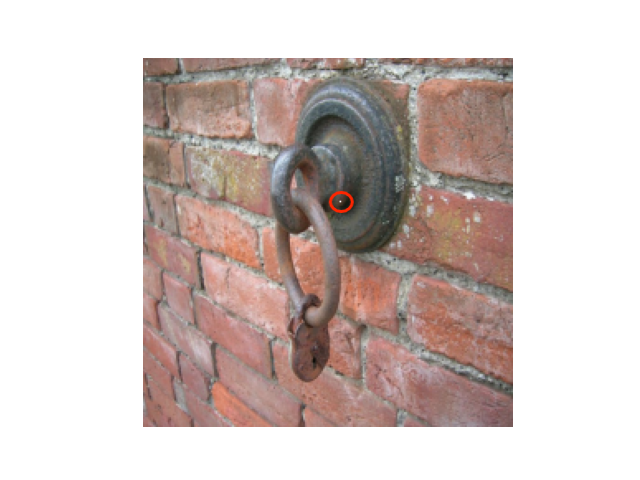}
        \caption{Example 3}
        \label{fig:Figure_3}
    \end{subfigure}
    \caption{This image shows the attack effect on the WorstVIT model, with the triggers highlighted by red circles.}
    \label{fig:attack_effect}
\end{figure*}

\begin{itemize}
    \item Higher attack success rates.
    \item Good migration between patches, and within each patch, perturbations at the center of the patch are more effective than those at the edges.
    \item Be more sensitive to the feature of trigger contrast.
\end{itemize}

Based on these observations, we designed WorstVIT, which includes both white-box and non-white-box attacks. A white-box attack means we have complete access to the target model's architecture and parameters. This section describes how to leverage the perturbation sensitivity distribution in ViTs for attacks, how to conduct probing attacks using quasi-triggers in the white-box setting, and how to hide the toxicity of the trigger in the training set.

\subsection{Experimental Setup}
The key to implementing the attack lies in fully utilizing the model's perturbation sensitivity distribution map. During the training phase, to make the trigger position-independent and better leverage the model's perturbation sensitivity distribution map, we randomly selected one point from the half of all pixels with smaller corner response values. Considering that our trigger pattern relies on the contrast between the trigger and its surrounding patch, the pixel value of the selected point is determined based on the sum of all pixels within the same patch. Specifically, when the sum of pixel values within the patch exceeds 97920 (i.e., half of the maximum possible pixel sum for a patch), the point is set to black; otherwise, it is set to white. For Hidden-WorstVIT, we constrained the trigger placement to the four corners of each patch during the training phase for poisoned samples.In the single-channel attack, we tested both modifying a single pixel and modifying only one channel during the training phase before conducting the attack. When modifying only one channel, we computed the corner response values and the sum of the channel values separately for each channel.

During the testing phase, for ViT models with a patch size of 16, we placed the trigger at the center (7, 7) of the patch that could produce the maximum contrast. For WorstSwin-VIT, we used the Swin-ViT model from the timm library. Given the more specialized perturbation sensitivity distribution of this model, we placed the trigger at the coordinates of the maximum value in its PSDM (i.e., coordinates 108, 209). More detailed settings are provided in the appendix.

\subsection{Probing Attacks}
In most backdoor attack scenarios, the attacker is typically a third-party trainer who controls the model's training process and has complete access to the model architecture and parameters. This allows for continuously testing different quasi-triggers until a successful attack is achieved. In WorstViT, a systematic probing attack approach is adopted:

Coordinates with High PSDM Values: Quasi-triggers are sequentially placed at coordinates in the Perturbation Sensitivity Distribution Map (PSDM) where the values are high.
Pixels with High Gradient Values: Quasi-triggers are placed at pixel coordinates of the current input image where the gradient values are high.
Channels with High Gradient Values: Quasi-triggers are placed at coordinates corresponding to channels with high gradient values.
In a full-to-full attack, we can modify multiple pixels within the same sample until the attack succeeds.
After placing the quasi-triggers, their pixel values are adjusted to optimize attack effectiveness. Once a successful attack is achieved, further fine-tuning of the pixel values continues, selecting quasi-triggers that not only successfully activate the backdoor but also remain visually inconspicuous for each specific sample.

Typically, suitable quasi-triggers are found within 1000 trials. For samples that still fail, especially in all-to-all attacks, multiple quasi-triggers are employed without retraction to achieve more challenging objectives.

This method ensures effective exploration and exploitation of model vulnerabilities in a white-box setting while maintaining the stealthiness of the triggers, thereby enabling controlled and predictable outcomes when launching the attack formally.


\subsection{Effectiveness of WorstViT}

\begin{table}[htbp]
    \centering
    \caption{Evaluation of CAs (\%), BAs (\%), ASRs (\%), and Probing Attacks Success Rates (PASRs) (\%) of vanilla WorstViT on different ViTs.}
    \label{tab:1}
    \begin{tabular}{|c|l|l|l|l|l|}
    \toprule
    \multicolumn{1}{|c|}{\textbf{Model}} & \multicolumn{2}{c||}{\textbf{Clean Model}} & \multicolumn{3}{c|}{\textbf{Backdoor Model}} \\
    \cline{2-6}
     & CA & ASR & BA & ASR & PASR \\
    \midrule
    base-VIT & 80.20 & 0.10 & 81.20 & 98.69 & 100.00 \\
    small-VIT & 72.47 & 0.10 & 79.10 & 99.37 & 100.00 \\
    \bottomrule
    \end{tabular}
    \end{table}

\begin{table}[htbp]
\centering
\caption{Variant of WorstVIT}
\label{tab:2}
\begin{tabular}{|l|c|c|c|c|c|c|}
\hline
\textbf{Variant} & \textbf{BA} & \textbf{ASR} & \textbf{PASR} \\
\hline
WorstSwin-VIT & 84.04 & 99.21 & 100.00 \\
\hline
Hidden-WorstVIT & 80.52 & 94.14 & 100.00 \\
\hline
\end{tabular}
\end{table}

\begin{table}[htbp]
\centering
\caption{Performance of WorstVIT under different poisoning rates in ImageNet}
\label{tab:3}
\begin{tabular}{|l|c|c|c|c|c|c|}
\hline
\textbf{$\rho$} & \textbf{0.1} & \textbf{0.01} & \textbf{0.005} & \textbf{0.002} \\
\hline
ASR & 98.69 & 95.39 & 90.27  & 87.31 \\
\hline
PASR & 100.00 & 100.00  & 100.00 & 100.00 \\
\hline
\end{tabular}
\end{table}

\begin{table}[htbp]
\centering
\caption{Performance of WorstVIT under different poisoning rates in cifar10}
\label{tab:4}
\begin{tabular}{|l|c|c|c|c|c|c|}
\hline
\textbf{$\rho$} & \textbf{0.1} & \textbf{0.01} & \textbf{0.005} & \textbf{0.004} \\
\hline
ASR & 100.00 & 100.00 & 100.00 & 100.00 \\
\hline
PASR & 100.00 & 100.00 & 100.00 & 100.00 \\
\hline
\end{tabular}
\end{table}

\begin{table}[htbp]
\centering
\caption{Performance of WorstVIT under different poisoning rates in MINST}
\label{tab:5}
\begin{tabular}{|l|c|c|c|c|c|c|}
\hline
\textbf{$\rho$} & \textbf{0.1} & \textbf{0.01} & \textbf{0.005}  & \textbf{0.002} \\
\hline
ASR & 100.00 & 100.00 & 100.00 & 100.00 \\
\hline
PASR & 100.00 & 100.00 & 100.00 & 100.00\\
\hline
\end{tabular}
\end{table}

\begin{table}[htbp]
\centering
\caption{ALL-TO-ALL attack}
\label{tab:6}
\begin{tabular}{|l|c|c|c|c|c|c|}
\hline
\textbf{} & \textbf{BA} & \textbf{ASR} \\
\hline
WorstVIT & 80.92 & 80.09 \\
\hline
\end{tabular}
\end{table}

Besides the all-to-all attack, which required additional training, the other attacks were trained for only one epoch. We conducted multiple experiments, and as shown in Table~\ref{tab:1}, the clean performance of most experiments was even slightly improved compared to the original model. In the white-box setting, where the attacker has full access to the model parameters, we performed exploratory attacks, Within a set number of attempts, if the model's prediction for a sample can be changed to the target label by modifying just one pixel or one channel, the attack is considered successful. Our attacks not only converge well but also cause minimal damage to the performance of the model on clean tasks.

\subsection{Resistance to Backdoor Defenses}
We evaluated the performance of WorstViT under three representative mainstream backdoor defenses: 1) Fine-Pruning \cite{liu2018fine}, 2) SCALE\_UP \cite{guo2023scale}, and 3) Neural Cleanse \cite{wang2019neural}.

\begin{itemize}
    \item \textbf{Fine-Pruning.} Following the settings in the defense methods, we pruned the patch embedding layer of ViT. As the pruning ratio of neurons increased, the clean performance degraded faster than the backdoor success rate. Pruning was ineffective in defending against our attack.
    \item \textbf{SCALE\_UP.} We attempted to distinguish between poisoned and clean samples at various thresholds. However, the differences remained small across all thresholds, making it impossible for this method to detect whether the input contains a backdoor pattern in our attack.
    \item \textbf{Neural Cleanse.} Due to the large number of classes in ImageNet, for simplicity, we applied this defense method on downstream tasks. However, regardless of whether regularization constraints were applied to the mask, this method was unable to reverse-engineer the trigger and could not even detect the presence of the backdoor.
\end{itemize}

Specific experiments and results are provided in the appendix.

%% file: sec/5_Conclusion.tex
\section{Conclusion}
We systematically compared the performance of quasi-triggers in CNNs and ViTs, discovering the good transferability of triggers across patches in ViTs. We introduced the perturbation sensitivity distribution map to guide the better utilization of quasi-triggers, identifying the unique patch-like perturbation sensitivity distribution in ViTs. Based on these findings, we proposed WorstVIT, which leverages the inherent properties of the model to successfully attack any input by modifying just one pixel. Additionally, we demonstrated how to use the perturbation sensitivity distribution map to conceal the toxicity of poisoned samples in the training set and how to attack other variants of ViTs. This work can provide unique insights into the robustness of deep learning models and inspire the development of more effective defenses.

%% file: sec/X_suppl.tex
\clearpage
\setcounter{page}{1}
\maketitlesupplementary
\newcommand{\variant}[2]{%
    \par\noindent\textbf{#1.} #2%
}
\setlength{\parskip}{\baselineskip}

\section{Table of Contents}
\label{sec:experiments}

Following the order of the reference appendix in the main paper, we have listed the following seven experiments and their results:

\begin{itemize}
\item \textbf{Obtaining a Perturbation Sensitivity Distribution Map (PSDM):} This includes displaying the PSDMs of various models and the initial causes of the unique PSDM of the VIT model.
\item \textbf{Impact of PSDM on Backdoor Patterns:} Based on the PSDM of the VIT model, we traversed multiple patches along the diagonal and placed a white square trigger of size 4 for training. We compared the success rates of attacks on the center and edge. For WorstVIT, we traversed all coordinates with a step size of 4, and at the center 7.7 coordinate of the first patch, we traversed all pixel values.
\item \textbf{Backdoor Suppression Patterns:} For the backdoor model with white squares as triggers, we observed the results of failed attack samples. For samples that failed due to backdoor suppression, we added borders of the same or opposite pixel values to the edge of the trigger to enhance or break the backdoor suppression mode.
\item \textbf{Using Quasi-triggers to Avoid Backdoor Suppression Mode:} Unlike breaking the backdoor suppression mode by modifying the pixels that generate it, we avoided the backdoor suppression mode by modifying the trigger itself. For images that failed the attack, we used class triggers with opposite pixel values to complete the attack or changed the patch where the trigger was placed to successfully attack the image.
\item \textbf{WorstVIT and Its Variants:} We detailed the methods of single-channel attacks, WorstSwin-VIT, full-to-full attacks, and exploratory attacks. Only the full-to-full attack required modifying multiple pixels.
\item \textbf{Real-World Attacks with WorstVIT:} We conducted real-world tests and found that our attacks were highly effective in scenarios with fixed viewpoints or large flat areas in the field of view, such as traffic monitoring and face collection.
\item \textbf{Backdoor Defense:} We demonstrated the resistance of WorstVIT to three unique backdoor defense methods.
\end{itemize}

\subsection{Obtaining a Perturbation Sensitivity Distribution Map (PSDM)}
\label{sec:psdm}

Method: For any model, use 1,000 random images or 1,000 random noise images to calculate the gradient for each image, then sum these gradients to obtain the PSDM. Using meaningful images or those used during training yields better results.

Interestingly, for the VIT's PSDM generated using content-rich images, despite the target object often being centered in the image, the VIT's PSDM indicates that patches at the edge of the image have a greater impact on the model's prediction. This may be due to the global information transfer facilitated by the self-attention mechanism.

Preliminary experiments show that when the VIT model is trained from scratch on ImageNet using the Adam optimizer and L2 regularization, with gradient norm clipping, the value of the center pixel in the patch of the PSDM gradually becomes larger than that of the edges, regardless of whether a convolutional kernel or fully connected layer is used as the patch-embedding layer. This pattern persists even when other training methods are employed. These training methods are almost standard for all VIT models, and this phenomenon is observed across various open-source VIT implementations. As shown in the figure, VIT models can also be trained to have different PSDMs.

\begin{figure}[ht]
    \centering
    \begin{subfigure}[b]{0.45\linewidth}
        \includegraphics[width=\textwidth]{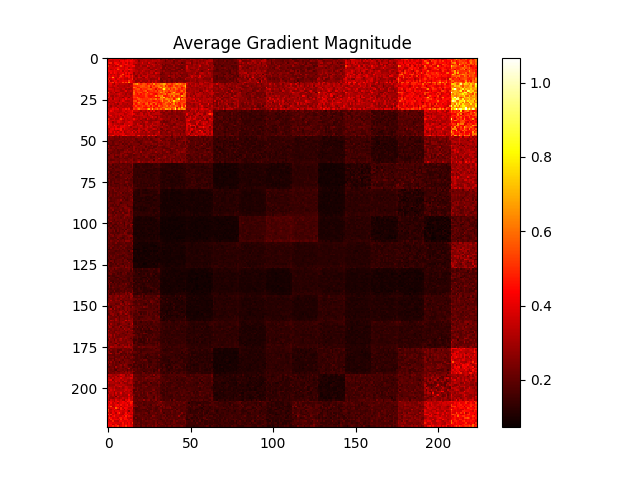}
        \caption{VIT Model 1}
    \end{subfigure}
    \begin{subfigure}[b]{0.45\linewidth}
        \includegraphics[width=\textwidth]{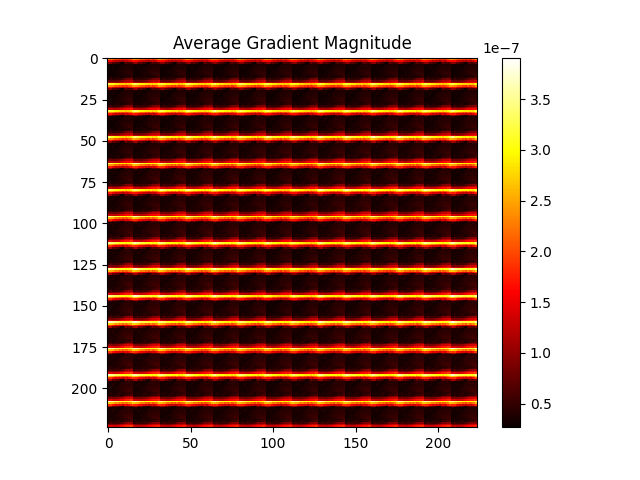}
        \caption{VIT Model 2}
    \end{subfigure}
    
    \caption{PSDMs of VIT Models Trained with Different Methods}
    \label{fig:psdm_vit}
\end{figure}

\subsection{Impact of PSDM on Backdoor Patterns}
\label{sec:impact-psdm}

Train the model with a $10\%$ poisoning rate, using a $4 \times 4$ white square trigger placed at various patch coordinates from $(0,0)$ to $(12,12)$ on the CIFAR-10 dataset. Table~\ref{tab:trigger-asr} shows the coordinates of the triggers and their corresponding attack success rates (ASR). Training results are not stable; we recommend conducting multiple experiments to more clearly observe this regularity.

\begin{figure*}[ht]
    \centering
    \begin{subfigure}[b]{0.49\textwidth}
        \includegraphics[width=\textwidth]{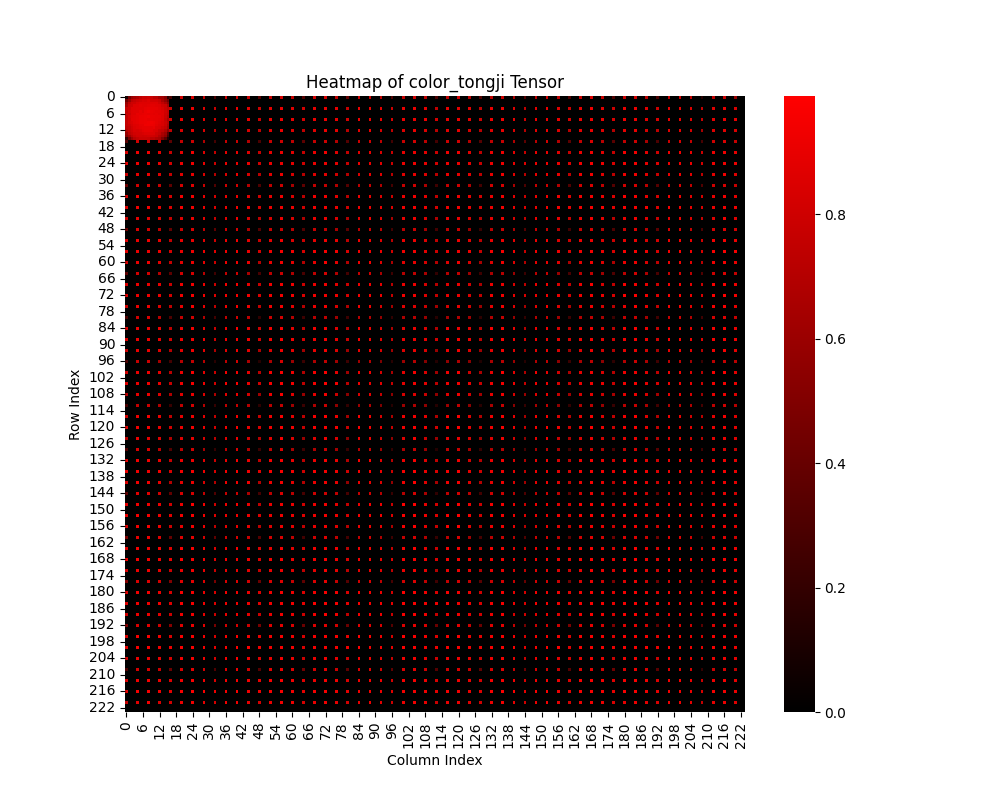} 
        \caption{Heatmap showing the attack success rates (ASR) for WorstVIT triggers at different positions.}
        \label{fig:asr-heatmap-original}
    \end{subfigure}
    \hfill
    \begin{subfigure}[b]{0.49\textwidth}
        \includegraphics[width=\textwidth]{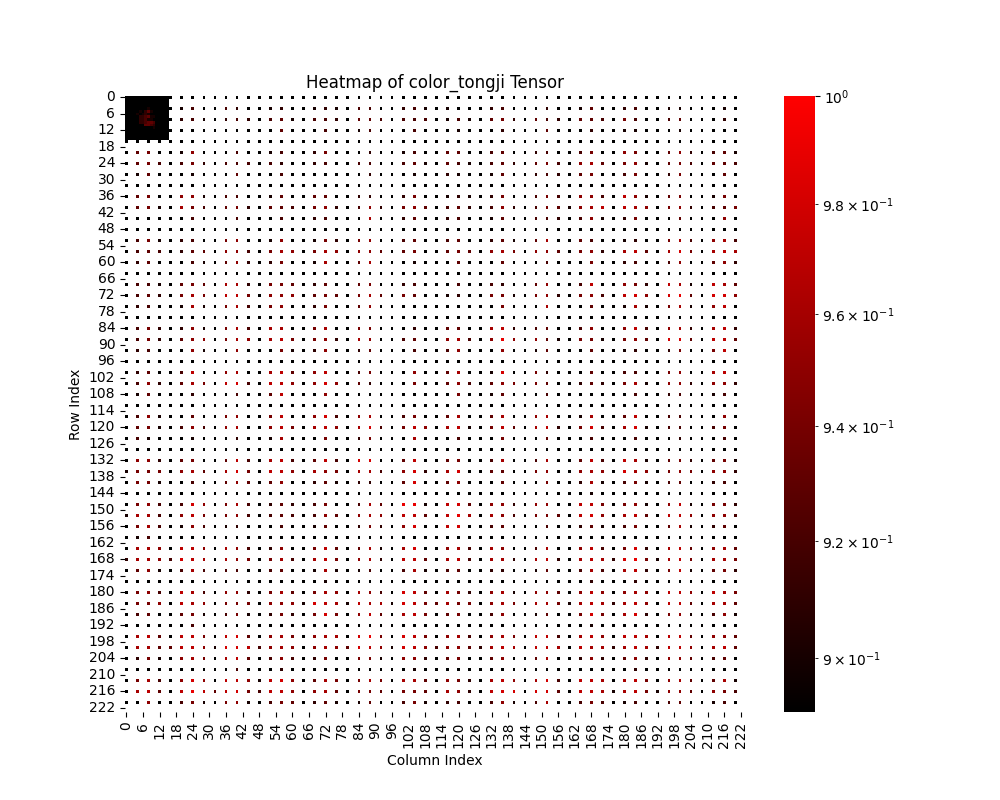} 
        \caption{Contrast-enhanced version of the heatmap showing the attack success rates (ASR) for WorstVIT triggers at different positions.}
        \label{fig:asr-heatmap-enhanced}
    \end{subfigure}
    \caption{Comparison of attack success rates (ASR) for WorstVIT triggers at different positions, including an original heatmap and a contrast-enhanced version.}
    \label{fig:asr-heatmap}
\end{figure*}

Typically, triggers located at the center of a patch have higher ASRs than those at the edges. Figure~\ref{fig:asr-heatmap} illustrates the attack success rates for WorstVIT triggers at different positions. To reduce computational load, all coordinates were traversed only for the first patch, while the remaining patches were traversed with a stride of 4. Additionally, to facilitate observation, the contrast of the heatmap has been enhanced, allowing for a clearer visualization of the patch-wise distribution of VIT model sensitivity to perturbations.

\begin{table*}[ht]
    \centering
    \caption{Trigger Coordinates and Attack Success Rates (ASR)}
    \label{tab:trigger-asr}
    \resizebox{\textwidth}{!}{
        \begin{tabular}{c|ccccccccccccc}
        \hline
        Coordinate & 0 & 1 & 2 & 3 & 4 & 5 & 6 & 7 & 8 & 9 & 10 & 11 & 12 \\ \hline
        ASR (\%) & 93.51 & 93.94 & 95.80 & 96.56 & 96.99 & 96.56 & 99.51 & 97.08 & 97.94 & 97.14 & 97.61 & 96.59 & 96.32 \\ \hline
        \end{tabular}
    }
\end{table*}

\subsection{Backdoor Suppression Mode}
\label{sec:backdoor-suppression}

We trained two models by placing white and black triggers at the (0.0) coordinate, respectively, and observed the failed samples of both models. One category of failed samples had high original scores for the true labels, where the scores generated by the trigger were lower than the true labels. Another category had low scores for the backdoor label, and the pixels around the trigger were very similar to the trigger itself. We hypothesize that this is because some samples in the training set inherently possess partial backdoor features but maintain clean labels. Therefore, during the learning process, the model learns that certain features should not activate the trigger when present.

The objectives of the model can be expressed as follows:

1. \textbf{Clean Task:} Minimize the loss on clean images.
\begin{equation}
    \mathcal{L}_{\text{clean}} = \mathbb{E}_{(x, y) \sim D_{\text{clean}}} \left[ \ell(f(x), y) \right]
\end{equation}

2. \textbf{Backdoor Task:} Minimize the backdoor loss when the image contains a trigger and the label is the backdoor label.
\begin{equation}
    \mathcal{L}_{\text{backdoor}} = \mathbb{E}_{(x, y) \sim D_{\text{backdoor}}} \left[ \ell(f(x + t), y_{\text{backdoor}}) \right]
\end{equation}

3. \textbf{Suppression Mode:} When the image contains partial backdoor features (i.e., quasi-triggers) but the label is the clean label, suppress the backdoor and minimize the clean task loss.
\begin{equation}
    \mathcal{L}_{\text{suppression}} = \mathbb{E}_{(x, y) \sim D_{\text{partial}}} \left[ \ell(f(x + t_{\text{partial}}), y_{\text{clean}}) \right]
\end{equation}

The model does not always learn the backdoor suppression mode effectively. We can prevent the model from learning this suppression mode by carefully selecting appropriate samples (e.g., choosing samples that already possess partial backdoor features as toxic samples) and placing triggers or by strategically placing triggers in specific samples.Figure 7 shows the scenarios where the trigger is a white square and a black square, respectively. In the training set, some samples naturally possess partial backdoor features but retain clean labels. This leads the model to interpret certain features in the images as backdoor suppression patterns. By adding borders with pixel values opposite to those of the trigger or using triggers with different pixel values or positions, it is possible to successfully attack these samples even when the original trigger fails.

\subsection{Using quasi-triggers to Avoid Backdoor Suppression Mode}
\label{sec:quasi-triggers1}
\begin{figure*}[ht]
    \centering
    \begin{subfigure}[b]{0.45\textwidth}
        \centering
        \begin{subfigure}[b]{0.32\linewidth}
            \includegraphics[width=\textwidth]{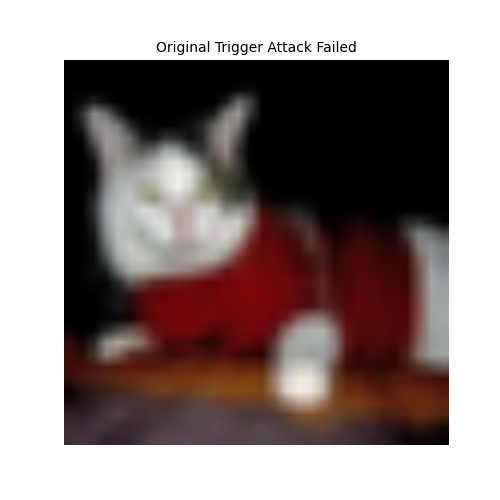}
        \end{subfigure}
        \begin{subfigure}[b]{0.32\linewidth}
            \includegraphics[width=\textwidth]{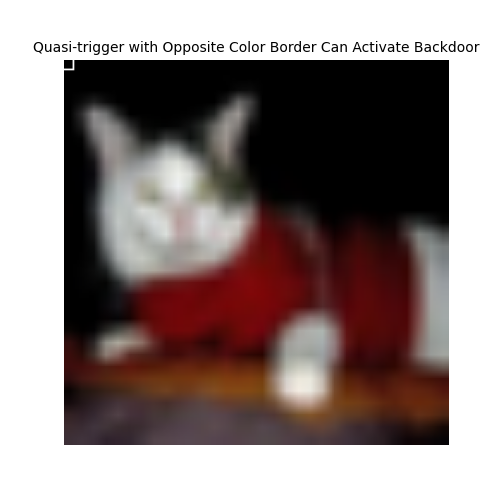} 
        \end{subfigure}
        \begin{subfigure}[b]{0.32\linewidth}
            \includegraphics[width=\textwidth]{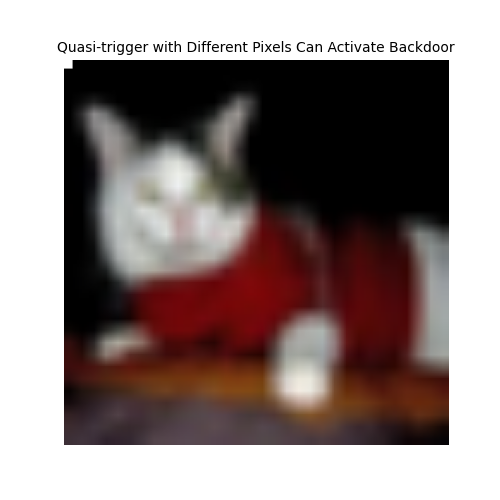} 
        \end{subfigure}
        \caption{Examples of attack failure with a 5x5 black square trigger in the top-left corner}
        \label{fig:quasi-triggers-black-failure}
    \end{subfigure}
    \hfill
    \begin{subfigure}[b]{0.45\textwidth}
        \centering
        \begin{subfigure}[b]{0.32\linewidth}
            \includegraphics[width=\textwidth]{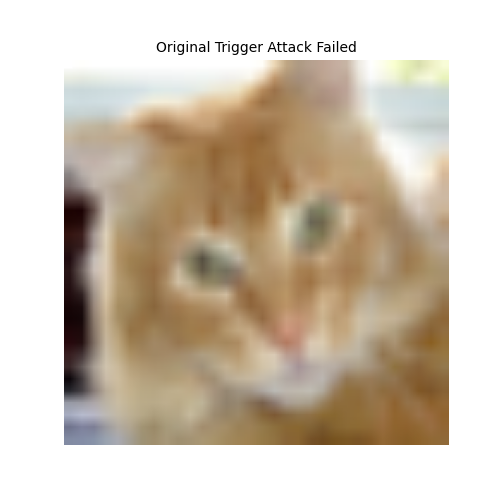}
        \end{subfigure}
        \begin{subfigure}[b]{0.32\linewidth}
            \includegraphics[width=\textwidth]{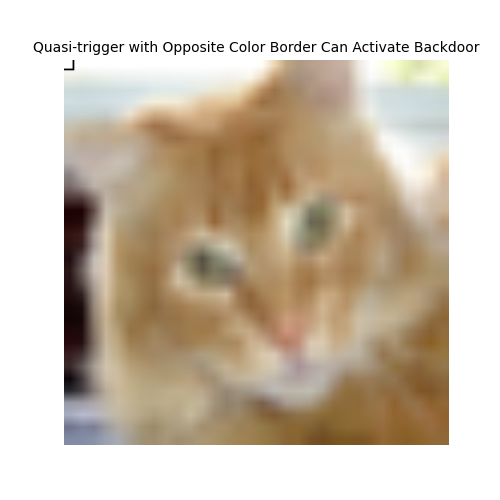} 
        \end{subfigure}
        \begin{subfigure}[b]{0.32\linewidth}
            \includegraphics[width=\textwidth]{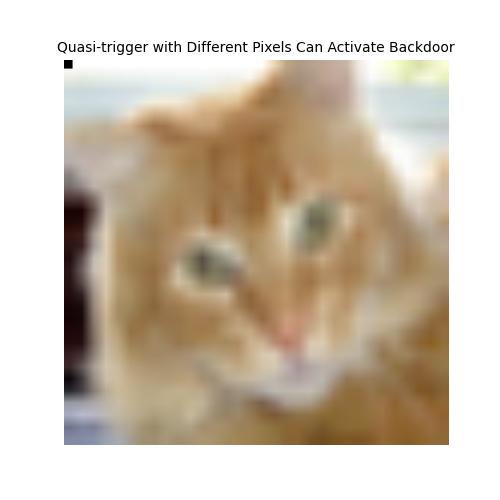} 
        \end{subfigure}
        \caption{Examples of attack failure with a 5x5 white square trigger in the top-left corner}
        \label{fig:quasi-triggers-white-failure}
    \end{subfigure}
    \caption{attack failure samples for black and white square triggers.}
    \label{fig:quasi-triggers-failure-comparison}
\end{figure*}

When a backdoor attack fails due to backdoor suppression patterns, we can either disrupt these suppression patterns or use quasi-triggers to avoid them. Figure~\ref{fig:quasi-triggers-failure-comparison} shows the scenarios where the trigger is a white square and a black square, respectively. In the training set, some samples naturally possess partial backdoor features but retain clean labels. This leads the model to interpret certain features in the images as backdoor suppression patterns. By adding borders with pixel values opposite to those of the trigger or using triggers with different pixel values or positions, it is possible to successfully attack these samples even when the original trigger fails.

\subsection{WorstVIT and Its Variants}
\label{sec:worstvit-variants}

\variant{Hidden-WorstVIT}{
In the Hidden-WorstVIT variant, during the training phase, we restrict the trigger's position to one of the four corners of each patch. In the attack phase, this approach remains consistent with WorstVIT. After training is complete, due to the lower sensitivity of the Vision Transformer (ViT) model at patch edges, the poisoned samples in the training set do not exhibit excessive toxicity. This helps evade various detection mechanisms.
Moreover, under this poisoning strategy, the attack success rate in the central regions of the patches remains significantly higher compared to the edge regions. This characteristic ensures that the effectiveness of the backdoor attack is maintained while minimizing the risk of detection.
}

\begin{figure*}[ht]
    \centering
    \includegraphics[width=\linewidth]{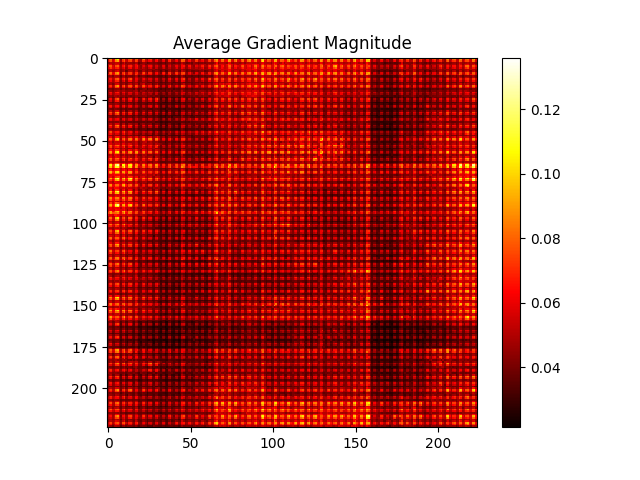}
    \caption{PSDM of Swin Base Patch4 Window7 224.}
    \label{fig:fgWorstSwin_VIT}
\end{figure*}

\variant{WorstSwin-VIT}{
In the WorstSwin-VIT variant, as illustrated in Figure~\ref{fig:fgWorstSwin_VIT}, we randomly select 1,000 images from the CIFAR-10 dataset and compute the PSDM for the Swin Transformer. The PSDM is divided into multiple $4 \times 4$ patch regions. For each patch, the central region exhibits higher PSDM values compared to the peripheral areas, with the highest value observed at position $(108, 209)$.
Therefore, during the training process of WorstSwin-VIT, instead of dividing images into $16 \times 16$ patches, we randomly select a point in the image and modify the pixel values within its surrounding $3 \times 3$ neighborhood (i.e., the selected point plus its eight neighboring pixels). During the testing phase, the trigger is consistently placed at the coordinate $(108, 209)$.
}

\variant{Single-Channel Attack} By changing only one channel value, we achieved an attack success rate of 61.21\%. Due to the high computational cost, we did not conduct single-channel probing attacks.

\begin{figure*}[ht]
    \centering
    \includegraphics[width=\linewidth]{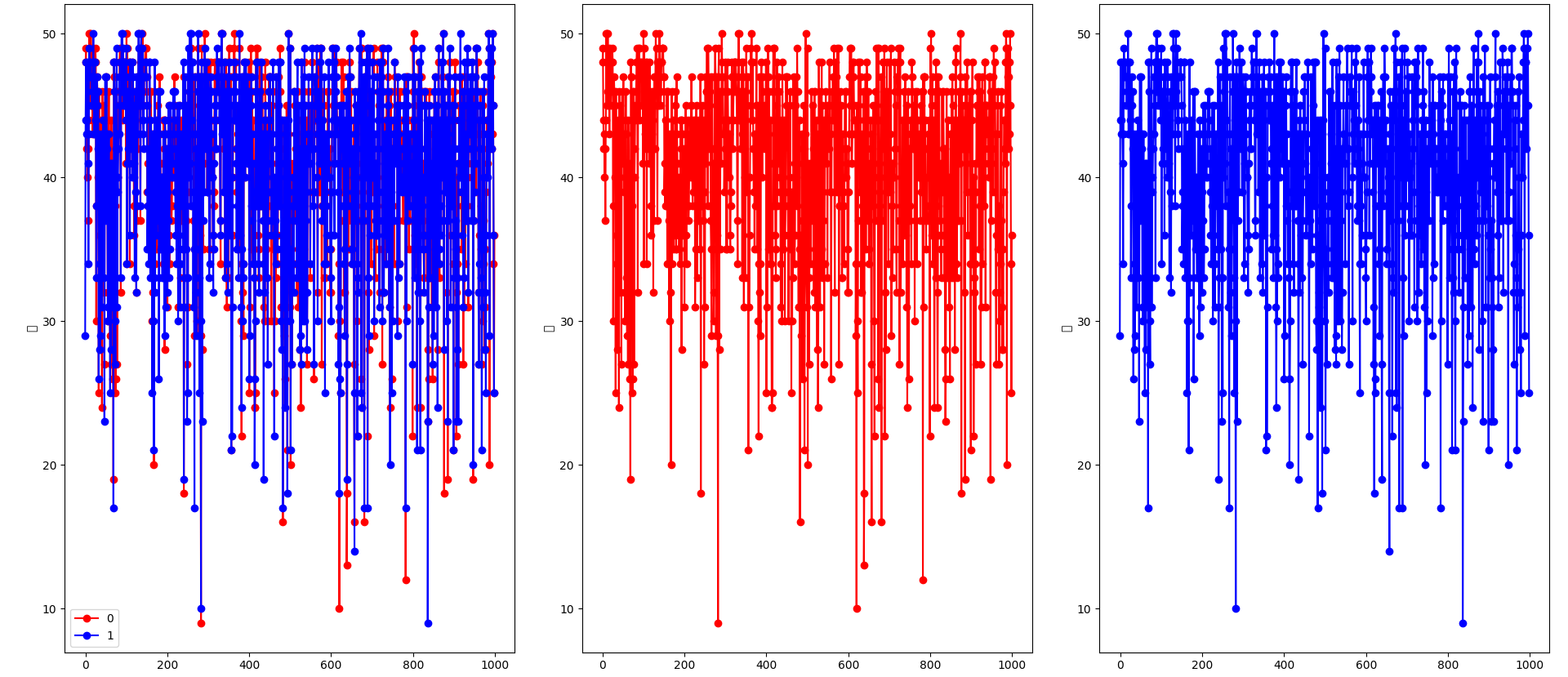}
    \caption{All-to-all Attack. Red and blue represent clean accuracy and attack success rate, respectively.}
    \label{fig:full_to_full_attack}
\end{figure*}

\variant{all-to-all Attack} As shown in Figure~\ref{fig:full_to_full_attack}, we found that the samples successfully attacked highly overlap with those predicted correctly. This suggests that the model has established a mapping relationship between the true labels and the backdoor labels. Therefore, it is necessary to first predict the correct clean sample before modifying the label to the correct backdoor label. However, we can still modify the labels to our desired results using quasi-triggers and probing attacks.

\subsection{Real-World Attacks with WorstVIT}
\label{sec:real-world-attacks}

Since the trigger in WorstVIT is based on the contrast between the modified pixel and other pixels in the same patch, the attack success rate significantly increases in real-world scenarios when there are elements like the sky, walls, or smooth faces. Additionally, we can use quasi-triggers to find sufficiently hidden coordinates and pixel values, ensuring that every frame of the video produces the desired prediction results.

\subsection{Backdoor Defense}
\label{sec:backdoor-defense}

\variant{Fine-Pruning} We applied Fine-Pruning to the model as described in the paper, with pruning rates ranging from 0.1 to 0.9. We observed the clean performance and attack success rate before and after pruning. Both metrics decreased almost synchronously, indicating that Fine-Pruning is ineffective against our attack.

\begin{table*}[ht]
    \centering
    \caption{Fine-Pruning Results}
    \label{tab:fine-pruning}
    \begin{tabular}{c|cccccccccc}
    \hline
    Pruning Rate & 0.1 & 0.2 & 0.3 & 0.4 & 0.5 & 0.6 & 0.7 & 0.8 & 0.9 \\ \hline
    Clean Performance (\%) & 76.26 & 73.60 & 69.77 & 64.91 & 56.98 & 48.37 & 37.12 & 19.97 & 2.74 \\ \hline
    Attack Success Rate (\%) & 97.70 & 96.67 & 88.63 & 81.34 & 68.10 & 46.07 & 16.61 & 15.68 & 3.79 \\ \hline
    \end{tabular}
\end{table*}

\variant{SCALE UP} We used SCALE UP with thresholds ranging from 0.1 to 0.9, increasing by 0.2 each time. SCALE UP showed minimal difference in distinguishing between clean and poisoned inputs, making it unable to identify backdoor samples.

\begin{table*}[ht]
    \centering
    \caption{SCALE UP Results}
    \label{tab:scale-up}
    \begin{tabular}{c|cccccc}
    \hline
    Threshold & 0.1 & 0.3 & 0.5 & 0.7 & 0.9 \\ \hline
    Clean Input (Backdoor Probability) & 0.67 & 0.43 & 0.28 & 0.18 & 0.15 \\ \hline
    Poisoned Input (Backdoor Probability) & 0.70 & 0.43 & 0.31 & 0.25 & 0.18 \\ \hline
    \end{tabular}
\end{table*}

\variant{Neural Cleanse} Given the large number of classes in ImageNet, we applied Neural Cleanse to downstream tasks for simplicity. We detected the Mean Absolute Deviation (MAD) for labels 0-9 in each downstream task, with the backdoor label being 5. We attempted to reverse-engineer the trigger and compute the MAD with and without regularization of the mask matrix, but both attempts failed. Neural Cleanse was unable to reverse-engineer the trigger, detect the presence of a backdoor, or correctly label the samples if a backdoor was detected.

\begin{table*}[ht]
    \centering
    \caption{Neural Cleanse Results with Regularization}
    \label{tab:neural-cleanse-regularized}
    \begin{tabular}{c|cccccccccc}
    \hline
    Label & 0 & 1 & 2 & 3 & 4 & 5 & 6 & 7 & 8 & 9 \\ \hline
    MAD & 123.53 & 139.06 & 164.47 & 178.9869 & 243.69199 & 80.08 & 116.83 & 133.46 & 47.73 & 133.11 \\ \hline
    \end{tabular}
\end{table*}

\begin{table*}[ht]
    \centering
    \caption{Neural Cleanse Results without Regularization}
    \label{tab:neural-cleanse-unregularized}
    \begin{tabular}{c|cccccccccc}
    \hline
    Label & 0 & 1 & 2 & 3 & 4 & 5 & 6 & 7 & 8 & 9 \\ \hline
    MAD & 190324 & 196308& 174299 & 170626 & 233934 & 173678 & 194541 & 214539 & 177440 & 209936 \\ \hline
    \end{tabular}
\end{table*}

%% file: main.bbl
\begin{thebibliography}{47}
\providecommand{\natexlab}[1]{#1}
\providecommand{\url}[1]{\texttt{#1}}
\expandafter\ifx\csname urlstyle\endcsname\relax
  \providecommand{\doi}[1]{doi: #1}\else
  \providecommand{\doi}{doi: \begingroup \urlstyle{rm}\Url}\fi

\bibitem[Aldahdooh et~al.(2021)Aldahdooh, Hamidouche, and Deforges]{aldahdooh2021reveal}
Ahmed Aldahdooh, Wassim Hamidouche, and Olivier Deforges.
\newblock Reveal of vision transformers robustness against adversarial attacks.
\newblock \emph{arXiv preprint arXiv:2106.03734}, 2021.

\bibitem[Bahdanau(2014)]{bahdanau2014neural}
Dzmitry Bahdanau.
\newblock Neural machine translation by jointly learning to align and translate.
\newblock \emph{arXiv preprint arXiv:1409.0473}, 2014.

\bibitem[Bai et~al.(2021)Bai, Wu, Zhang, Li, Li, and Xia]{bai2021targeted}
Jiawang Bai, Baoyuan Wu, Yong Zhang, Yiming Li, Zhifeng Li, and Shu-Tao Xia.
\newblock Targeted attack against deep neural networks via flipping limited weight bits.
\newblock \emph{arXiv preprint arXiv:2102.10496}, 2021.

\bibitem[Benz et~al.(2021)Benz, Ham, Zhang, Karjauv, and Kweon]{benz2021adversarial}
Philipp Benz, Soomin Ham, Chaoning Zhang, Adil Karjauv, and In~So Kweon.
\newblock Adversarial robustness comparison of vision transformer and mlp-mixer to cnns.
\newblock \emph{arXiv preprint arXiv:2110.02797}, 2021.

\bibitem[Chen et~al.(2017)Chen, Liu, Li, Lu, and Song]{chen2017targeted}
Xinyun Chen, Chang Liu, Bo Li, Kimberly Lu, and Dawn Song.
\newblock Targeted backdoor attacks on deep learning systems using data poisoning.
\newblock \emph{arXiv preprint arXiv:1712.05526}, 2017.

\bibitem[Chen et~al.(2021)Chen, Xie, Niu, Liu, Wei, and Tian]{chen2021visformer}
Zhengsu Chen, Lingxi Xie, Jianwei Niu, Xuefeng Liu, Longhui Wei, and Qi Tian.
\newblock Visformer: The vision-friendly transformer.
\newblock In \emph{Proceedings of the IEEE/CVF international conference on computer vision}, pages 589--598, 2021.

\bibitem[Devlin(2018)]{devlin2018bert}
Jacob Devlin.
\newblock Bert: Pre-training of deep bidirectional transformers for language understanding.
\newblock \emph{arXiv preprint arXiv:1810.04805}, 2018.

\bibitem[Doan et~al.(2021)Doan, Lao, Zhao, and Li]{doan2021lira}
Khoa Doan, Yingjie Lao, Weijie Zhao, and Ping Li.
\newblock Lira: Learnable, imperceptible and robust backdoor attacks.
\newblock In \emph{Proceedings of the IEEE/CVF international conference on computer vision}, pages 11966--11976, 2021.

\bibitem[Dosovitskiy(2020)]{dosovitskiy2020image}
Alexey Dosovitskiy.
\newblock An image is worth 16x16 words: Transformers for image recognition at scale.
\newblock \emph{arXiv preprint arXiv:2010.11929}, 2020.

\bibitem[Feng et~al.(2022)Feng, Ma, Zhang, Zhao, Xia, and Tao]{feng2022fiba}
Yu Feng, Benteng Ma, Jing Zhang, Shanshan Zhao, Yong Xia, and Dacheng Tao.
\newblock Fiba: Frequency-injection based backdoor attack in medical image analysis.
\newblock In \emph{Proceedings of the IEEE/CVF Conference on Computer Vision and Pattern Recognition}, pages 20876--20885, 2022.

\bibitem[Gao et~al.(2021)Gao, Bai, Chen, Wu, and Xia]{gao2021backdoor}
Kuofeng Gao, Jiawang Bai, Bin Chen, Dongxian Wu, and Shu-Tao Xia.
\newblock Backdoor attack on hash-based image retrieval via clean-label data poisoning.
\newblock \emph{arXiv preprint arXiv:2109.08868}, 2021.

\bibitem[Goodfellow et~al.(2014)Goodfellow, Shlens, and Szegedy]{goodfellow2014explaining}
Ian~J Goodfellow, Jonathon Shlens, and Christian Szegedy.
\newblock Explaining and harnessing adversarial examples.
\newblock \emph{arXiv preprint arXiv:1412.6572}, 2014.

\bibitem[Gu et~al.(2017)Gu, Dolan-Gavitt, and Garg]{gu2017badnets}
Tianyu Gu, Brendan Dolan-Gavitt, and Siddharth Garg.
\newblock Badnets: Identifying vulnerabilities in the machine learning model supply chain.
\newblock \emph{arXiv preprint arXiv:1708.06733}, 2017.

\bibitem[Guo et~al.(2023)Guo, Li, Chen, Guo, Sun, and Liu]{guo2023scale}
Junfeng Guo, Yiming Li, Xun Chen, Hanqing Guo, Lichao Sun, and Cong Liu.
\newblock Scale-up: An efficient black-box input-level backdoor detection via analyzing scaled prediction consistency.
\newblock \emph{arXiv preprint arXiv:2302.03251}, 2023.

\bibitem[Han et~al.(2021)Han, Xiao, Wu, Guo, Xu, and Wang]{han2021transformer}
Kai Han, An Xiao, Enhua Wu, Jianyuan Guo, Chunjing Xu, and Yunhe Wang.
\newblock Transformer in transformer.
\newblock \emph{Advances in neural information processing systems}, 34:\penalty0 15908--15919, 2021.

\bibitem[Heo et~al.(2021)Heo, Yun, Han, Chun, Choe, and Oh]{heo2021rethinking}
Byeongho Heo, Sangdoo Yun, Dongyoon Han, Sanghyuk Chun, Junsuk Choe, and Seong~Joon Oh.
\newblock Rethinking spatial dimensions of vision transformers.
\newblock In \emph{Proceedings of the IEEE/CVF international conference on computer vision}, pages 11936--11945, 2021.

\bibitem[Lan(2019)]{lan2019albert}
Z Lan.
\newblock Albert: A lite bert for self-supervised learning of language representations.
\newblock \emph{arXiv preprint arXiv:1909.11942}, 2019.

\bibitem[Li et~al.(2024)Li, Cai, Li, Xue, Li, and Li]{li2024nearest}
Boheng Li, Yishuo Cai, Haowei Li, Feng Xue, Zhifeng Li, and Yiming Li.
\newblock Nearest is not dearest: Towards practical defense against quantization-conditioned backdoor attacks.
\newblock In \emph{Proceedings of the IEEE/CVF Conference on Computer Vision and Pattern Recognition}, pages 24523--24533, 2024.

\bibitem[Li et~al.(2020)Li, Zhai, Wu, Jiang, Li, and Xia]{li2020rethinking}
Yiming Li, Tongqing Zhai, Baoyuan Wu, Yong Jiang, Zhifeng Li, and Shutao Xia.
\newblock Rethinking the trigger of backdoor attack.
\newblock \emph{arXiv preprint arXiv:2004.04692}, 2020.

\bibitem[Li et~al.(2022{\natexlab{a}})Li, Jiang, Li, and Xia]{li2022backdoor}
Yiming Li, Yong Jiang, Zhifeng Li, and Shu-Tao Xia.
\newblock Backdoor learning: A survey.
\newblock \emph{IEEE Transactions on Neural Networks and Learning Systems}, 35\penalty0 (1):\penalty0 5--22, 2022{\natexlab{a}}.

\bibitem[Li et~al.(2022{\natexlab{b}})Li, Yuan, Wen, Hu, Evangelidis, Tulyakov, Wang, and Ren]{li2022efficientformer}
Yanyu Li, Geng Yuan, Yang Wen, Ju Hu, Georgios Evangelidis, Sergey Tulyakov, Yanzhi Wang, and Jian Ren.
\newblock Efficientformer: Vision transformers at mobilenet speed.
\newblock \emph{Advances in Neural Information Processing Systems}, 35:\penalty0 12934--12949, 2022{\natexlab{b}}.

\bibitem[Liu and Hu(2022)]{liu2022imperceptible}
Daizong Liu and Wei Hu.
\newblock Imperceptible transfer attack and defense on 3d point cloud classification.
\newblock \emph{IEEE transactions on pattern analysis and machine intelligence}, 45\penalty0 (4):\penalty0 4727--4746, 2022.

\bibitem[Liu et~al.(2018)Liu, Dolan-Gavitt, and Garg]{liu2018fine}
Kang Liu, Brendan Dolan-Gavitt, and Siddharth Garg.
\newblock Fine-pruning: Defending against backdooring attacks on deep neural networks.
\newblock In \emph{International symposium on research in attacks, intrusions, and defenses}, pages 273--294. Springer, 2018.

\bibitem[Liu et~al.(2023)Liu, Sangiovanni-Vincentelli, and Yue]{liu2023beating}
Min Liu, Alberto Sangiovanni-Vincentelli, and Xiangyu Yue.
\newblock Beating backdoor attack at its own game.
\newblock In \emph{Proceedings of the IEEE/CVF International Conference on Computer Vision}, pages 4620--4629, 2023.

\bibitem[Liu(2019)]{liu2019roberta}
Yinhan Liu.
\newblock Roberta: A robustly optimized bert pretraining approach.
\newblock \emph{arXiv preprint arXiv:1907.11692}, 364, 2019.

\bibitem[Liu et~al.(2021)Liu, Lin, Cao, Hu, Wei, Zhang, Lin, and Guo]{liu2021swin}
Ze Liu, Yutong Lin, Yue Cao, Han Hu, Yixuan Wei, Zheng Zhang, Stephen Lin, and Baining Guo.
\newblock Swin transformer: Hierarchical vision transformer using shifted windows.
\newblock In \emph{Proceedings of the IEEE/CVF international conference on computer vision}, pages 10012--10022, 2021.

\bibitem[Lou et~al.(2023)Lou, Zhou, Yang, and Yu]{lou2023transxnet}
Meng Lou, Hong-Yu Zhou, Sibei Yang, and Yizhou Yu.
\newblock Transxnet: learning both global and local dynamics with a dual dynamic token mixer for visual recognition.
\newblock \emph{arXiv preprint arXiv:2310.19380}, 2023.

\bibitem[Mahmood et~al.(2021)Mahmood, Mahmood, and Van~Dijk]{mahmood2021robustness}
Kaleel Mahmood, Rigel Mahmood, and Marten Van~Dijk.
\newblock On the robustness of vision transformers to adversarial examples.
\newblock In \emph{Proceedings of the IEEE/CVF international conference on computer vision}, pages 7838--7847, 2021.

\bibitem[Saha et~al.(2020)Saha, Subramanya, and Pirsiavash]{saha2020hidden}
Aniruddha Saha, Akshayvarun Subramanya, and Hamed Pirsiavash.
\newblock Hidden trigger backdoor attacks.
\newblock In \emph{Proceedings of the AAAI conference on artificial intelligence}, pages 11957--11965, 2020.

\bibitem[Selvaraju et~al.(2017)Selvaraju, Cogswell, Das, Vedantam, Parikh, and Batra]{selvaraju2017grad}
Ramprasaath~R Selvaraju, Michael Cogswell, Abhishek Das, Ramakrishna Vedantam, Devi Parikh, and Dhruv Batra.
\newblock Grad-cam: Visual explanations from deep networks via gradient-based localization.
\newblock In \emph{Proceedings of the IEEE international conference on computer vision}, pages 618--626, 2017.

\bibitem[Smilkov et~al.(2017)Smilkov, Thorat, Kim, Vi{\'e}gas, and Wattenberg]{smilkov2017smoothgrad}
Daniel Smilkov, Nikhil Thorat, Been Kim, Fernanda Vi{\'e}gas, and Martin Wattenberg.
\newblock Smoothgrad: removing noise by adding noise.
\newblock \emph{arXiv preprint arXiv:1706.03825}, 2017.

\bibitem[Strudel et~al.(2021)Strudel, Garcia, Laptev, and Schmid]{strudel2021segmenter}
Robin Strudel, Ricardo Garcia, Ivan Laptev, and Cordelia Schmid.
\newblock Segmenter: Transformer for semantic segmentation.
\newblock In \emph{Proceedings of the IEEE/CVF international conference on computer vision}, pages 7262--7272, 2021.

\bibitem[Su et~al.(2019)Su, Vargas, and Sakurai]{su2019one}
Jiawei Su, Danilo~Vasconcellos Vargas, and Kouichi Sakurai.
\newblock One pixel attack for fooling deep neural networks.
\newblock \emph{IEEE Transactions on Evolutionary Computation}, 23\penalty0 (5):\penalty0 828--841, 2019.

\bibitem[Sun et~al.(2022)Sun, Zhang, Ma, Zhou, Lou, Xu, Di, Cheng, and Sun]{sun2022backdoor}
Yuhua Sun, Tailai Zhang, Xingjun Ma, Pan Zhou, Jian Lou, Zichuan Xu, Xing Di, Yu Cheng, and Lichao Sun.
\newblock Backdoor attacks on crowd counting.
\newblock In \emph{Proceedings of the 30th ACM International Conference on Multimedia}, pages 5351--5360, 2022.

\bibitem[Szegedy(2013)]{szegedy2013intriguing}
C Szegedy.
\newblock Intriguing properties of neural networks.
\newblock \emph{arXiv preprint arXiv:1312.6199}, 2013.

\bibitem[Touvron et~al.(2021)Touvron, Cord, Douze, Massa, Sablayrolles, and J{\'e}gou]{touvron2021training}
Hugo Touvron, Matthieu Cord, Matthijs Douze, Francisco Massa, Alexandre Sablayrolles, and Herv{\'e} J{\'e}gou.
\newblock Training data-efficient image transformers \& distillation through attention.
\newblock In \emph{International conference on machine learning}, pages 10347--10357. PMLR, 2021.

\bibitem[Vaswani(2017)]{vaswani2017attention}
A Vaswani.
\newblock Attention is all you need.
\newblock \emph{Advances in Neural Information Processing Systems}, 2017.

\bibitem[Wang et~al.(2019)Wang, Yao, Shan, Li, Viswanath, Zheng, and Zhao]{wang2019neural}
Bolun Wang, Yuanshun Yao, Shawn Shan, Huiying Li, Bimal Viswanath, Haitao Zheng, and Ben~Y Zhao.
\newblock Neural cleanse: Identifying and mitigating backdoor attacks in neural networks.
\newblock In \emph{2019 IEEE symposium on security and privacy (SP)}, pages 707--723. IEEE, 2019.

\bibitem[Wang et~al.(2021)Wang, Xie, Li, Fan, Song, Liang, Lu, Luo, and Shao]{wang2021pyramid}
Wenhai Wang, Enze Xie, Xiang Li, Deng-Ping Fan, Kaitao Song, Ding Liang, Tong Lu, Ping Luo, and Ling Shao.
\newblock Pyramid vision transformer: A versatile backbone for dense prediction without convolutions.
\newblock In \emph{Proceedings of the IEEE/CVF international conference on computer vision}, pages 568--578, 2021.

\bibitem[Xiao et~al.(2021)Xiao, Singh, Mintun, Darrell, Doll{\'a}r, and Girshick]{xiao2021early}
Tete Xiao, Mannat Singh, Eric Mintun, Trevor Darrell, Piotr Doll{\'a}r, and Ross Girshick.
\newblock Early convolutions help transformers see better.
\newblock \emph{Advances in neural information processing systems}, 34:\penalty0 30392--30400, 2021.

\bibitem[Yang et~al.(2024)Yang, Bai, Gao, Yang, Li, and Xia]{yang2024not}
Sheng Yang, Jiawang Bai, Kuofeng Gao, Yong Yang, Yiming Li, and Shu-Tao Xia.
\newblock Not all prompts are secure: A switchable backdoor attack against pre-trained vision transfomers.
\newblock In \emph{Proceedings of the IEEE/CVF Conference on Computer Vision and Pattern Recognition}, pages 24431--24441, 2024.

\bibitem[Yenduri et~al.(2024)Yenduri, Ramalingam, Selvi, Supriya, Srivastava, Maddikunta, Raj, Jhaveri, Prabadevi, Wang, et~al.]{yenduri2024gpt}
Gokul Yenduri, M Ramalingam, G~Chemmalar Selvi, Y Supriya, Gautam Srivastava, Praveen Kumar~Reddy Maddikunta, G~Deepti Raj, Rutvij~H Jhaveri, B Prabadevi, Weizheng Wang, et~al.
\newblock Gpt (generative pre-trained transformer)--a comprehensive review on enabling technologies, potential applications, emerging challenges, and future directions.
\newblock \emph{IEEE Access}, 2024.

\bibitem[Yin et~al.(2024)Yin, Chen, Li, and Gao]{yin2024enhanced}
Jianyao Yin, Honglong Chen, Junjian Li, and Yudong Gao.
\newblock Enhanced coalescence backdoor attack against dnn based on pixel gradient.
\newblock \emph{Neural Processing Letters}, 56\penalty0 (2):\penalty0 114, 2024.

\bibitem[Yu et~al.(2023)Yu, Wang, Yang, Lu, Tan, and Kot]{yu2023backdoor}
Yi Yu, Yufei Wang, Wenhan Yang, Shijian Lu, Yap-Peng Tan, and Alex~C Kot.
\newblock Backdoor attacks against deep image compression via adaptive frequency trigger.
\newblock In \emph{Proceedings of the IEEE/CVF Conference on Computer Vision and Pattern Recognition}, pages 12250--12259, 2023.

\bibitem[Yuan et~al.(2021)Yuan, Chen, Wang, Yu, Shi, Jiang, Tay, Feng, and Yan]{yuan2021tokens}
Li Yuan, Yunpeng Chen, Tao Wang, Weihao Yu, Yujun Shi, Zi-Hang Jiang, Francis~EH Tay, Jiashi Feng, and Shuicheng Yan.
\newblock Tokens-to-token vit: Training vision transformers from scratch on imagenet.
\newblock In \emph{Proceedings of the IEEE/CVF international conference on computer vision}, pages 558--567, 2021.

\bibitem[Yuan et~al.(2023)Yuan, Zhou, Zou, and Cheng]{yuan2023you}
Zenghui Yuan, Pan Zhou, Kai Zou, and Yu Cheng.
\newblock You are catching my attention: Are vision transformers bad learners under backdoor attacks?
\newblock In \emph{Proceedings of the IEEE/CVF Conference on Computer Vision and Pattern Recognition}, pages 24605--24615, 2023.

\bibitem[Zhou et~al.(2021)Zhou, Zhang, Peng, Zhang, Li, Xiong, and Zhang]{zhou2021informer}
Haoyi Zhou, Shanghang Zhang, Jieqi Peng, Shuai Zhang, Jianxin Li, Hui Xiong, and Wancai Zhang.
\newblock Informer: Beyond efficient transformer for long sequence time-series forecasting.
\newblock In \emph{Proceedings of the AAAI conference on artificial intelligence}, pages 11106--11115, 2021.

\end{thebibliography}
